\def\year{2020}\relax
\documentclass[letterpaper]{article} 
\usepackage{aaai20}  
\usepackage{times}  
\usepackage{helvet} 
\usepackage{courier}  
\usepackage[hyphens]{url}  
\usepackage{graphicx} 
\usepackage{graphicx}  
\usepackage{amsmath}
\usepackage{amssymb}
\usepackage{subfigure}
\usepackage{epstopdf}
\usepackage{algorithm}
\usepackage{algpseudocode}
\usepackage{multirow}
\usepackage{array}
\usepackage[titletoc,toc,page]{appendix}

\title{Efficient Querying from Weighted Binary Codes}
\author{\Large \textbf{Zhenyu Weng, Yuesheng Zhu}\\
Communication and Information Security Laboratory, Shenzhen Graduate School, Peking University\\
wzytumbler@pku.edu.cn, zhuys@pku.edu.cn
}
 
\begin{document}

\maketitle

\begin{abstract}
Binary codes are widely used to represent the data due to their small storage and efficient computation. However, there exists an ambiguity problem that lots of binary codes share the same Hamming distance to a query. To alleviate the ambiguity problem, weighted binary codes assign different weights to each bit of binary codes and compare the binary codes by the weighted Hamming distance. Till now, performing the querying from the weighted binary codes efficiently is still an open issue. In this paper, we propose a new method to rank the weighted binary codes and return the nearest weighted binary codes of the query efficiently. In our method, based on the multi-index hash tables, two algorithms, the table bucket finding algorithm and the table merging algorithm, are proposed to select the nearest weighted binary codes of the query in a non-exhaustive and accurate way. The proposed algorithms are justified by proving their theoretic properties. The experiments on three large-scale datasets validate both the search efficiency and the search accuracy of our method. Especially for the number of weighted binary codes up to one billion, our method shows a great improvement of more than 1000 times faster than the linear scan.
\end{abstract}

\section{Introduction}
With the explosive growth of data, binary codes are widely used to represent the data due to their small storage and efficient computation. Given a query, the nearest binary codes can be ranked and returned efficiently by computing the Hamming distance between the query and the binary codes. BRISK~\cite{6126542}, ORB~\cite{6126544}, and other binary image descriptors~\cite{7882718} are designed to represent the image data, and successfully used in various applications, including image matching, 3D reconstruction and object recognition. End-to-end feature learning methods~\cite{DBLP:conf/aaai/LiDWXL19,DBLP:conf/aaai/SongHGXHS18} based on the neural networks extract the binary codes from the images and are widely used in image retrieval and cross retrieval. In addition to these specific image binary codes, hashing methods~\cite{liu2014discrete,DBLP:conf/aaai/LinJLSWW19,DBLP:journals/pami/LiuJWS19,7915742} are used to map different high-dimensional feature vectors into compact binary codes. Since these feature vectors may represent image data, video data, or other multimedia data, the hashing methods can be used in various multimedia retrieval applications.

However, the number of possible Hamming distance is limited and different binary codes may share the same Hamming distance to the given binary query. To alleviate this ambiguity problem and further improve the performance of binary codes, weighted binary codes are used~\cite{Fan2013Learning,gordo2014asymmetric,zhang2013binary}. By assigning bitwise weights to each bit of binary codes, the distance between a pair of binary codes is calculated by weighted Hamming distance instead of Hamming distance. For example, ~\cite{Huang2017RWBD,Fan2013Learning} are designed to learn the weights for the binary image descriptors to improve their discriminative power for image matching. And ~\cite{Duan2015Weighted,weng2016asymmetric} are designed to learn the weights for the binary codes generated by different hashing methods to improve their search accuracy for multimedia retrieval.

Although weighted binary codes can alleviate the ambiguity problem, querying from the binary codes by weighted Hamming distance is slower than that by Hamming distance. To accelerate the querying process from the weighted binary codes, some methods~\cite{gordo2014asymmetric} use lookup tables to compute the query-independent values in advance. However, it is still an exhaustive linear scan. Some methods~\cite{Duan2015Weighted,norouzi2014fast} use Hamming distance to find the neighbors that have the smallest Hamming distance to the query and rank them according to the weighted Hamming distance. This non-exhaustive way is fast but cannot return the nearest weighted binary codes of the query accurately, resulting in a degraded performance of weighted binary codes in the application.

\begin{figure*}[t]
\centering
\includegraphics[width=0.96\textwidth]{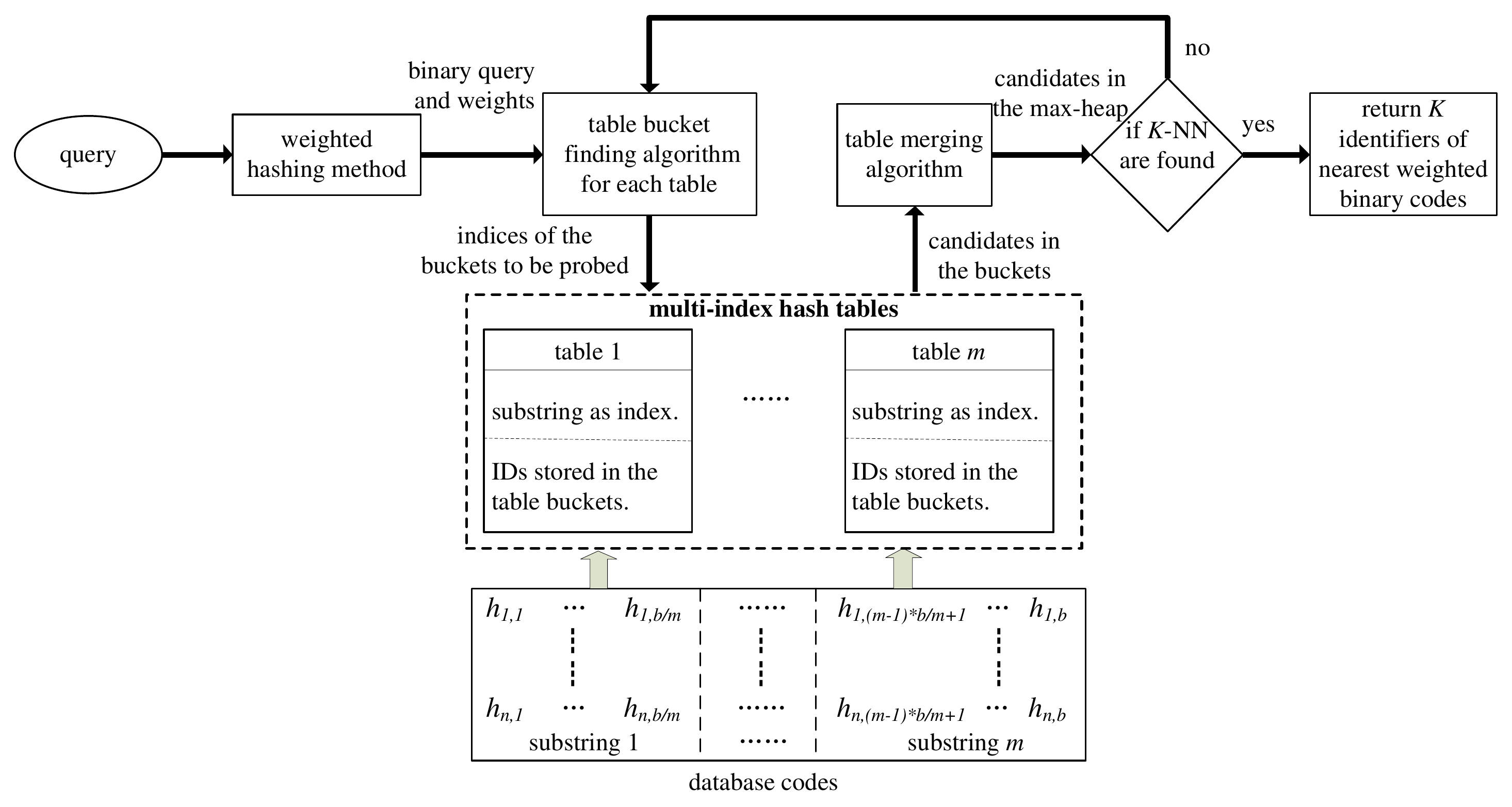}
\caption{The diagram of our method to find the $K$ nearest weighted binary codes of the query.}
\label{fig:pipeline}

\end{figure*}

In this paper, we propose a new method to rank the weighted binary codes and return the nearest weighted binary codes of the query in a non-exhaustive but accurate way. The diagram of our method is shown in Fig.~\ref{fig:pipeline}. Based on the multi-index hash tables~\cite{norouzi2014fast} on the binary code substrings, our method can efficiently choose the candidates in each table and merge the candidates to select the nearest weighted binary codes of the query. Theoretical analysis is provided to prove the our method can return the same ranking result as the linear scan does on the weighted binary codes. And the experiments show that our method is much faster than the linear scan.

\section{Related Work}
\subsection{Multi-Index Hash Tables on Binary Codes}
To avoid the exhaustive linear search on the binary codes, multi-index hash tables~\cite{norouzi2014fast} are built to accelerate the search on the binary codes and to return the $K$ nearest binary codes of the query in a non-exhaustive way.

In the multi-index tables~\cite{norouzi2014fast}, to index the binary codes from the database, $m$ different hash tables are built based on $m$ disjoint substrings of the binary codes as the index. If a binary code differs from the query by $r$ bits, it is an $r$-neighbor of the query. And the multi-index tables can find the $r$-neighbors of the query efficiently by probing each table. To return the $K$ nearest binary codes of the query, the Hamming search radius $r$ is progressively increased to find the $r$-neighbors of the query, until $K$ nearest binary codes are found.

In~\cite{norouzi2014fast}, the author mentioned that the multi-index tables can be used to return the top $K$ weighted binary codes by using Hamming distance to find the candidates that have the smallest Hamming distance to the query and culling them according to the weighted Hamming distance. However, this method cannot return the $K$ nearest weighted binary codes accurately. When increasing the search radius progressively until $K$ neighbors are found, it guarantees that the binary codes that are found have smaller Hamming distance to the query than the ones that are not found. In contrast, it cannot guarantee that these binary codes have the smaller weighted Hamming distance than the ones that are not found. The binary codes which have the larger Hamming distance from the query may have the smaller weighted Hamming distance.

\section{Querying from Weighted Binary Codes}
As shown in Fig.~\ref{alg1}, based on the multi-index hash tables, our method is composed of the table bucket finding algorithm and the table merging algorithm. Since we focus on finding the nearest neighbors of the query in the weighted Hamming space, in the following, we use the $K$-Nearest Neighbors ($K$-NN) of the query to denote the $K$ nearest weighted binary codes of the query.

\subsection{Table Bucket Finding Algorithm}

We start with a single-index hash table and propose a table bucket finding algorithm to find the table buckets in the single-index table. To further solve the long-code problem mentioned in~\cite{norouzi2014fast}, we extend to the multi-index hash tables and use the table bucket finding algorithm in each table. A table merging algorithm is proposed to merge the candidates from each table.

Assume a binary query $\bold{q} \in \{0,1\}^b$, a binary code $\bold{g} \in \{0,1\}^b$ and the weight functions $w_i(\cdot)$ for each bit are given, where $b$ is the length of the binary code and $w_i:\{0,1\}\to{\mathbb R}$. The weighted Hamming distance between the query $\bold{q}$ and the binary code $\bold{g}$ is defined as:
\begin{equation}
{d_w}({\bf{q}},{\bf{g}}) = \sum\limits_{i = 1}^b {{w_i} ({q_i} \oplus {g_i})},
\end{equation}
where $\oplus$ is an xor operation, $w_i(\cdot)$ is a weight function for the $i^{th}$ bit, $q_i$ is the $i^{th}$ bit of $\bold{q}$, and $g_i$ is the $i^{th}$ bit of $\bold{g}$.

Instead of finding the $K$-NN of the query $\bold{q}$ exhaustively, a single-index hash table is built by using the binary codes as the index of the hash table buckets. We probe the buckets in order from smallest to largest according to their weighted Hamming distance to the query, and take the identifiers as candidates in each probed table bucket until $K$ candidates are found. These $K$ candidates are the $K$-NN of the query.

The whole process of finding the buckets in order from smallest to largest can be regarded as multiple sequences combination problem (one bit represents one sequence). A algorithm~\cite{matsui2015pqtable,6915715} is used to solve the multiple sequences combination problem. However, this algorithm is not suitable in this situation. The algorithm can only traverse a few sequences (e.g. 2 or 4)  simultaneously to find the combination composing the bucket index that have the smallest weighted Hamming distance to the query. But in this situation, we have $b$ sequences where $b$ is much larger than 4 such that the traversal space is very large.

Based on the characteristic of the weighted binary codes, we propose a table bucket finding algorithm to find the bit combination which can compose the bucket index with the smallest weighted Hamming distance to the query. In the process of searching for the nearest neighbors of the query, since the query is fixed in each comparison between the query and the binary codes, the weight values for the xor result between the binary codes and the query can be pre-computed and stored. Hence, Eqn. (1) is rewritten as
\begin{equation}
{d_w}({\bf{g}}) = \sum\limits_{i = 1}^b {\hat w_i({g_i})},
\end{equation}
where $g_i$ is the $i^{th}$ bit of $\bold{g}$, $\hat w_i:\{0,1\}\to{\mathbb R}$ is a function to store the pre-computed weight value for the $i^{th}$ bit and is defined as:
\begin{equation}
\left\{ \begin{array}{l}
\hat w_i(0) = {w_i} (0 \oplus {q_i})\\
\hat w_i(1) = {w_i} (1 \oplus {q_i}).
\end{array} \right.
\end{equation}

As the input values of the function $\hat w_i(\cdot)$ are 0 or 1, correspondingly, there are two output values of $\hat w_i(\cdot)$. To construct a $b$-bit binary code $\bf{h}$$=[h_1\ldots h_b]$ that has the smallest weighted Hamming distance (smallest sum of weights) to the query, each bit $h_i$ of $\bold{h}$ is obtained as
\begin{equation}
{h_i} = \left\{ \begin{array}{l}
0\quad {\hat w_i}(0) \le {\hat w_i}(1)\\
1\quad otherwise.
\end{array} \right.
\end{equation}

When the $i^{th}$ bit of $\bf{h}$ is changed ($i.e.$ from 0 to 1 or from 1 to 0), we use $\bar h_i$ to denote the changed bit. When the bit is changed, the weight for this bit will increase. The increased weight $\Delta \hat w_i$ of the $i^{th}$ bit is defined as
\begin{equation}
\Delta \hat w_i = \hat w_i(\bar h_i) - \hat w_i({h_i}).
\end{equation}

The bits are ranked from smallest to largest according to $\Delta \hat w_i$ in advance. The leftmost bit has the smallest increased weight.

After ranking the bits and constructing the smallest binary code $\bold{h}$, to give the buckets to be probed in order from smallest to largest, we maintain a priority queue.  The top of the priority queue is the binary code that has the smallest sum of weights in the queue. $\bf{h}$ is the first one that is pushed into the priority queue. When taking out the top binary code $\bf{\tilde h}$ from the priority queue and probing the corresponding hash bucket, two new binary codes are constructed from $ \bold{\tilde h}$ by two different operations and pushed into the priority queue, respectively.

$\bold{Operation}$ $\bold{1}$ is to construct a binary code by changing the unchanged bit right next to the rightmost changed bit of $\bf{\tilde h}$ if the rightmost changed bit is not at the end of the current binary code. For example, assume ${\bf{\tilde h}} = [h_1 \ldots \bar h_r \ldots h_b]$, where $\bar h_r$ is rightmost changed bit. Then, the new binary code is constructed as ${\bf{\hat  h}} = [h_1 \ldots \bar h_r \bar h_{r+1} \ldots h_b]$.

$\bold{Operation}$ $\bold{2}$ is to construct a binary code by moving the rightmost changed bit of ${\bf{\tilde h}}$ to the next bit if the position of the rightmost changed bit is not at the end. For example, assume ${\bf{\tilde h}} = [h_1 \ldots \bar h_r \ldots h_b]$, where $\bar h_r$ is rightmost changed bit. Then, the new binary code is constructed as ${\bf{\dot  h}} = [h_1 \ldots h_r \bar h_{r+1} \ldots h_b]$.

For both operations, the new binary code has a larger sum of weights than the current one since the bits are ranked from smallest to largest according to Eqn.(5) in advance. It should be noted that for the initial binary code $\bf{h}$ which has no changed bit, only the first operation is permitted, which means to change the first bit of the binary code.

The pseudocode for querying with the single-index hash table is shown in Alg.~\ref{alg1}. Init() is a function that constructs the binary code $\bf{h}$ which has the smallest sum of weights according to the query $\bf{q}$ and the weights $\bold{w}$, and generates an order that denotes the positions of bits from smallest to largest according to Eqn.(5). Operation1() and Operation2() are two functions corresponding to above two operations to generate the new binary codes, respectively.

\begin{algorithm}
\caption{Querying with single-index hash table}               
\label{alg1}                      
\begin{algorithmic}[1]
\Require $\bold{q}$, $table$, $K$, weights $\bold{w}$              
\Ensure $u$    \Comment{a set of ranked identifiers}   
\State $u \leftarrow \emptyset$
\State $pri\_que \leftarrow \emptyset$    \Comment{priority queue}
\State $[pri\_que, order] \leftarrow$ Init($\bold{q}$, $\bold{w}$)
\While {$|u| < K$}
\State $code \leftarrow pri\_que.$top()
\State $pri\_que.$pop()    \Comment{remove top item from queue}
\State $pri\_que$.push(Operation1($code$, $order$))
\State $pri\_que$.push(Operation2($code$, $order$))
\State $ {\hat u}  \leftarrow table.$bucket($code$) \Comment{identifiers in the bucket}
\State $u.$extend($ {\hat u} $)
\EndWhile
\end{algorithmic}
\end{algorithm}

To prove that our algorithm can always find the binary code that has the smallest sum of weights among the un-probed binary codes, we begin with the following corollary.

$\bold{Corollary}$ $\bold{1:}$ Every binary code can be generated by above two operations.

$\bold{Proof.}$ A detailed proof is provided in Appendix A in the supplemental material.

We prove the correctness of our algorithm as follows:

$\bold{Theorem}$ $\bold{1:}$ The binary code that has the smallest sum of weights among the un-probed binary codes is always in the priority queue.

$\bold{Proof.}$ A detailed proof is provided in Appendix B in the supplemental material.

\subsection{Table Merging Algorithm}
As described in ~\cite{norouzi2014fast}, when the length of the binary code increases, the range of the table index expands and there are more table buckets in the table where a lot of buckets are empty. Traversing these empty table buckets is inefficient.

To solve this problem, following~\cite{norouzi2014fast}, $m$ different hash tables are built based on the $m$ disjoint substrings of the binary codes as the index in our method. The length of each substring is $\left\lceil {b/m} \right\rceil $ or $\left\lfloor {b/m} \right\rfloor $. For convenience, we assume that $b$ can be divided by $m$, and that the substrings comprise continuous bits.

Before describing the following table merging algorithm, we define $f:\{0,1\}^{b/m}\to{\mathbb R}$ as a function to calculate the sum of weights of the substring $\bold{s}$ in the table, which is:
\begin{equation}
f(\bold{s}) = \sum\limits_{j = 1}^{b/m} {{\hat w_j}({s_j})},
\end{equation}
where $s_j$ is the $j^{th}$ bit of $\bold{s}$ and ${\hat w_j}$ is the weight function for the $j^{th}$ bit in the corresponding table.

When given a query, each table maintains a priority queue according to the sum of weights of the corresponding substring. The priority queues operate the same as in Alg. 1 to find the un-probed bucket which is indexed by the corresponding substring and has the smallest sum of weights. Then, in each round we take out the top substring of each priority queue. By treating the substring as the index of the hash table bucket in the corresponding table, we can probe the table buckets and take the identifiers in each bucket as the candidates.

To merge the candidates from each bucket and determine if the $K$-NN of the query are found, a $K$-size max-heap is built to filter the candidates. The root node of the max-heap has the largest sum of weights in the heap. Assume the node $\bold{r} \in \{0,1\}^b$ in the max-heap is in the form of $\bold{r} = [\bold{r}_1, \ldots, \bold{r}_m]$ where $\bold{r}_i$ is the substring of $\bold{r}$ in the $i^{th}$ table. And a function $g:\{0,1\}^{b}\to{\mathbb R}$ to calculate the sum of weights of the node is defined as
\begin{equation}
g(\bold{r}) = \sum\limits_{i = 1}^m {f(\bold{r}_i)}.
\end{equation}

For each round, when the identifiers are taken from each table, they are compared to the root node in the max-heap. If an identifier $\bold{\hat r}$ has a smaller sum of weights than the root node $\bold{r}$ ($i.e.$ $g(\bold{\hat r}) < g(\bold{r})$), the root node is thrown away and the identifier is inserted into the max-heap. The process continues for multiple rounds until the root node of the max-heap is smaller or equal to a threshold.

In detail, assume there are $m$ tables and a $b$-bit binary code $\bold{h}$ is partitioned into $m$ disjoint substrings $\bold{s}$. When the top substring $\bold{s}_i$ of the $i^{th}$ priority queue is taken out, the queue will have the new top substring $\bold{\hat s}_i$. The associated identifiers from the table bucket $\bold{s}_i$ of the $i^{th}$ table are taken out and compared with the root node of the max-heap. The sum of weights from the top substring of each current priority queue is calculated as
\begin{equation}
S = \sum\limits_{i = 1}^m {f(\bold{\hat s}_i)}.
\end{equation}

If the max-heap has $K$ nodes and the root node $\bold{r}$ of the max-heap is smaller or equal to the sum of weights from the top substring of each current priority queue ($i.e.$ $g(\bold{r}) \le S$), the top $K$ nearest neighbors are found and the process stops.

We prove that the table merging algorithm can find the $K$-NN of the query with $\bold{Theorem}$ $\bold{2}$.

$\bold{Theorem}$ $\bold{2:}$ The binary codes that are found and stored in the max-heap have the smallest sum of weights among all binary codes.

$\bold{Proof}.$ A detailed proof is provided in Appendix C in the supplemental material.

We can further accelerate the searching process by reducing the number of the hash table buckets to be probed. In every round, assume the priority queues are ranked in some order $\bold{s}_{order}$. If the top substrings from the first $j$ queues are taken out, define the current sum of weights as below
\begin{equation}
{\hat S} = \sum\limits_{i = 1}^j {f(\bold{\hat s}_{order[i]}) + } \sum\limits_{i = j + 1}^m {f({\bold{s}_{order[i]}})}.
\end{equation}

Obviously, $\hat S \le S$ according to Eqn. (7). The searching process terminates when $g(\bold{r}) \le \hat S$. It has been proved that the binary codes that are found are the smallest among all the binary codes. From the equation, we can see that when the substring in the $i^{th}$ queue is taken out, $\hat S$ will increase $\Delta f_i$, which is defined as
\begin{equation}
\Delta f_i = f(\bold{\hat s}_i) - f({\bold{s}_i}).
\end{equation}

We want to make $\hat S$ smallest among all the orders such that the root node of the max-heap can be smaller or equal to $\hat S$ faster. Hence, the order can be obtained by ranking the priority queues from smallest to largest according to Eqn.(10).

The pseudocode for querying with multi-index hash tables is shown in Alg.~\ref{alg2}. Init(), Operation1() and Operation2() are the same functions as in the Alg.~\ref{alg1}. $m$ denotes the number of substrings for the binary code. $table[]$, $pri\_que[]$, $order[]$ denotes a set of tables, a set of priority queues, and a set of bit rankings for each substring, respectively. Split() is a function that splits the binary code and the weights into $m$ parts. $max\_heap$.satisfied() denotes whether max-heap has $K$ nodes and the root node of the max-heap satisfies the stopping criterion. $que\_order$ denotes the order of the priority queues to be checked. Sort() is a function that determines the ranking of the priority queues from smallest to largest according to Eqn.(10). As our method performs the querying process from the weighted binary codes based on the multi-index hash tables, we call it Multi-Index Weighted Querying (MIWQ).

\begin{algorithm}
\caption{Querying with multi-index hash tables}               
\label{alg2}                      
\begin{algorithmic}[1]
\Require $\bold{q}, table[], K, m$, weights $\bold{w}$              
\Ensure $max\_heap$     
\State $max\_heap \leftarrow \emptyset$
\State $[pri\_que[], order[]] \leftarrow$ Split(Init($\bold{q}$,$\bold{w}$),$m$)
\While {$!max\_heap.$satisfied()}
\For {$i \leftarrow$ $1$ to $m$}
\State $code[i] \leftarrow pri\_que[i].$top()
\State $pri\_que[i].$pop()
\State $pri\_que[i].$push(Operation1($code[i], order[i]$))
\State $pri\_que[i].$push(Operation2($code[i], order[i]$))
\EndFor
\State $que\_order \leftarrow$ Sort($pri\_que[]$.top(), $code[]$)
\For {$i \leftarrow$ $1$ to $m$}
\State $cur = que\_order[i]$
\State $max\_heap.$insert($table[cur].$hash($code[cur]$))
\If {$max\_heap.$satisfied()}
\State    break
\EndIf
\EndFor
\EndWhile
\end{algorithmic}
\end{algorithm}

\section{EXPERIMENTS}
\subsection{Datasets and Environment}

The experiments are performed on the three datasets: Places205, GIST1M and SIFT1B.

The Places205 dataset~\cite{zhou2014learning} is a scene-centric dataset with 205 scene categories. For each category, we randomly choose 5,000 images for search and 50 images as queries. Hence, we have 1,025,000 images for search and 10,250 queries. Each image is represented by a 128-D feature~\cite{Cakir2017MIHash}. The features are extracted from the fc7 layer of AlexNet~\cite{Krizhevsky2012ImageNet} pre-trained on ImageNet and reduced to 128 dimensions by PCA.

GIST1M dataset~\cite{jegou2011product} contains 1 million 960-D GIST descriptors~\cite{oliva2001modeling} which are global descriptors, and extracted from Tiny image set~\cite{torralba200880}. The dataset contains 1000 queries.

SIFT1B dataset~\cite{jegou2011product} contains 1 billion 128-D SIFT descriptors~\cite{lowe2004distinctive} and 10000 queries.

In the experiments, to evaluate the efficiency and the accuracy of different querying methods on the weighted binary codes, the classical data-independent hashing algorithm Locality-Sensitive Hashing (LSH)~\cite{andoni2006near} is used to map high-dimensional vectors into binary codes, and a weighted hashing method, Asymmetric Distance (Asym)~\cite{gordo2014asymmetric} is used to generate the weights for each bit of binary codes. All the experiments are run on a single core Intel Core-i7 CPU with 32GB of memory. The comparison of the querying methods on other binary codes and other weights is provided in Appendix D in the supplementary material to show the generality of our method.

\subsection{Comparison to Different Querying Methods}
Precision@$K$ is usually used to measure the accuracy of the approximate $K$-NN search.~\cite{7915742,matsui2015pqtable}. Here, we use precision@$K$ to evaluate whether our method can return the same results as the linear scan returns, and compare the performance of weighted binary codes in the approximate $K$-NN search with that of binary codes. The precision@$K$ is defined as the fraction of the true retrieved neighbors to the retrieved neighbors. It is formulated as follows
\begin{equation}
precision@K = \frac{{the\;true\;retrieved\;neighbors}}{K}
\end{equation}

For Places205, the ground truth refers to as the true neighbors the identifiers that have the same label as the query. For GIST1M and SIFT1B, the ground truth refers to as the true neighbors the top 1000 identifiers selected by linear scan with the Euclidean distance from the query in the original space, $i.e.$ Euclidean space.

The precision@$K$ results on Places205 and GIST1M are shown in Table~\ref{table:teaser1} and Table~\ref{table:teaser2}, respectively. In the tables, Baseline denotes querying from the binary codes according to Hamming distance, Linear Scan denotes querying from the binary code by the linear scan according to weighted Hamming distance, and Multi-Index Hashing (MIH)~\cite{norouzi2014fast} is a non-exhaustive but inexact querying method for the binary codes according to weighted Hamming distance. These querying methods are all implemented in C++. For MIH and our method, MIWQ, we use the same heuristic~\cite{norouzi2014fast} to determine the number of the substrings $m$, which is $b/log_2 n$ where $b$ is the length of the binary code and $n$ is the data size. According to the results, MIH can achieve higher search accuracy than Baseline, but is inferior to MIWQ. As MIH cannot return the $K$ nearest weighted binary codes accurately, MIH is inferior to MIWQ. Since MIWQ achieves the same search accuracy as Linear Scan, it shows that MIWQ can return the $K$ nearest weighted binary codes of the query accurately.

\begin{table}[!ht]
	\caption{The precision results on Places205.}
	\begin{center}
		\begin{tabular}
            {|c|c|c|c|c|}\hline
           \multirow{3}{*}{bit} & \multirow{3}{*}{method} & \multicolumn{3}{c|}{precision ($\%$)}                                 \\ \cline{3-5}
			                    &                         & 1-NN   & 10-NN    & 100-NN  \\ \hline
           \multirow{4}{*}{32}  &  Baseline               & 21.56  & 22.23	  &  19.83	\\ \cline{2-2}
                                &  MIH                    & 22.81  & 23.20	  &  20.86	 \\ \cline{2-2}
                                &  Linear Scan            & 25.29  & 24.97	  &  22.64	 \\ \cline{2-2}
                                & $\bold{MIWQ}$           & $\bf{25.29}$  & $\bf{24.97}$	  &  $\bf{22.64}$	\\ \hline
           \multirow{4}{*}{64}  &  Baseline               & 31.17  & 29.71	  &  26.67	\\ \cline{2-2}
                                &  MIH                    & 31.41  & 30.27	  &  27.46	\\ \cline{2-2}
                                &  Linear Scan            & 34.46  & 31.96	  &  28.72	\\ \cline{2-2}
                                & $\bold{MIWQ}$           & $\bf{34.46}$  &	$\bf{31.96}$  &	$\bf{28.72}$      \\ \hline
		\end{tabular}
	\end{center}
	\label{table:teaser1}
\end{table}

The speed-up factor is used to measure how fast our method and MIH are compared to the linear scan on the weighted binary codes. The speed-up factor is defined as dividing the run-time cost of the linear scan by the run-time cost of the test method, which is formulated as follows

\begin{equation}
speed{\rm{ - }}up\;factor = \frac{{time\;cost\;of\;linear\;scan}}{{time\;cost\;of\;test\;method}}
\end{equation}

\begin{table}[!ht]
	\caption{The precision results on GIST1M.}
	\begin{center}
		\begin{tabular}
            {|c|c|c|c|c|}\hline
           \multirow{3}{*}{bit} & \multirow{3}{*}{method} & \multicolumn{3}{c|}{precision ($\%$)}                                 \\ \cline{3-5}
			                    &                         & 1-NN   & 10-NN    & 100-NN   \\ \hline
           \multirow{4}{*}{32}  &  Baseline               & 11.90  & 7.57	  &  5.11	 \\ \cline{2-2}
                                &  MIH                    & 12.10  & 8.46	  &  5.92    \\ \cline{2-2}
                                &  Linear Scan            & 13.20  & 9.94     &	 7.31    \\ \cline{2-2}
                                & $\bold{MIWQ}$           & $\bf{13.20}$  &	$\bf{9.94}$ & $\bf{7.31}$ \\ \hline
           \multirow{4}{*}{64}  &  Baseline               & 21.10  & 14.47    &	 10.16	 \\ \cline{2-2}
                                &  MIH                    & 21.50  & 16.11    &	 11.55	 \\ \cline{2-2}
                                &  Linear Scan            & 24.90  & 19.14    &  13.65   \\ \cline{2-2}
                                & $\bold{MIWQ}$           & $\bf{24.90}$   & $\bf{19.14}$     &  $\bf{13.65}$    \\ \hline
		\end{tabular}
	\end{center}
	\label{table:teaser2}
\end{table}

\begin{table*}[!ht]
	\caption{The average time for the query on Places205.}
	\begin{center}
		\begin{tabular}
            {|c|c|c|c|c|c|c|c|}\hline
           \multirow{3}{*}{bit} & \multirow{3}{*}{method} & \multicolumn{6}{c|}{speed-up factors for $K$-NN}        \\ \cline{3-8}
		            &         &  \multicolumn{2}{c|}{1-NN} & \multicolumn{2}{c|}{10-NN} & \multicolumn{2}{c|}{100-NN} \\ \cline{3-8}
			               &         & time(ms) 	 & speed-up factor   & time(ms)  & speed-up factor  & time(ms)	& speed-up factor  \\ \hline
        \multirow{3}{*}{32} &  Linear Scan  &  23.06 & 1.0	& 23.06	 &  1.0	  &   23.06	 &  1.0   \\ \cline{2-2}
                                               &  MIH          &  0.11  & 209.6	& 0.21	 &  109.8 &   0.52	 &  44.3  \\ \cline{2-2}
                                               &  $\bf{MIWQ}$         &  $\bf{0.11}$  & $\bf{209.6}$	& $\bf{0.21}$	 & $\bf{109.8}$  &	  $\bf{0.55}$	 & $\bf{41.9}$   \\ \cline{1-8}
                             \multirow{3}{*}{64} &  Linear Scan  & 33.83  & 1.0	& 33.83	 & 1.0	  &   33.83	 & 1.0   \\ \cline{2-2}
                                                 &  MIH          &  0.51  & 66.3	& 1.03	 & 32.84  &   2.4	 & 14.0  \\ \cline{2-2}
                                                &  $\bf{MIWQ}$         &  $\bf{0.63}$  & $\bf{53.6}$	& $\bf{1.42}$	 & $\bf{23.82}$  &   $\bf{3.65}$	 & $\bf{9.2}$  \\ \hline

		\end{tabular}
	\end{center}
	\label{table:teaser3}
\end{table*}

\begin{table*}[!ht]
	\caption{The average time for the query on GIST1M.}
	\begin{center}
		\begin{tabular}
            {|c|c|c|c|c|c|c|c|}\hline
          \multirow{3}{*}{bit} & \multirow{3}{*}{method} & \multicolumn{6}{c|}{speed-up factors for $K$-NN}        \\ \cline{3-8}
		               &         &  \multicolumn{2}{c|}{1-NN} & \multicolumn{2}{c|}{10-NN} & \multicolumn{2}{c|}{100-NN} \\ \cline{3-8}
			                        &         & time(ms) 	 & speed-up factor    & time(ms)  & speed-up factor  & time(ms)	& speed-up factor  \\ \hline
         \multirow{3}{*}{32} &  Linear Scan  &  22.17 & 1.0	& 22.17	 & 1.0	  &   22.17	 & 1.0   \\ \cline{2-2}
                                                 &  MIH          &  0.1   & 221.7	& 0.16	 & 138.5  &   0.37	 & 59.9  \\ \cline{2-2}
                                                 &  $\bf{MIWQ}$         &  $\bf{0.12}$  & $\bf{184.7}$	& $\bf{0.25}$	 & $\bf{88.6}$	  &   $\bf{0.77}$	 & $\bf{28.7}$   \\ \cline{1-8}
                             \multirow{3}{*}{64} &  Linear Scan  & 39.36  & 1.0	& 39.36	 & 1.0	  &   39.36	 & 1.0   \\ \cline{2-2}
                                                 &  MIH          &  0.87  & 45.2	& 1.65	 & 23.8	  &   3.15	 & 12.4  \\ \cline{2-2}
                                                 &  $\bf{MIWQ}$         &  $\bf{1.84}$  & $\bf{21.3}$	& $\bf{3.9}$	 & $\bf{10.0}$	  &   $\bf{8.17}$	 & $\bf{4.8}$  \\ \hline

		\end{tabular}
	\end{center}
	\label{table:teaser4}
\end{table*}

Table~\ref{table:teaser3} and Table~\ref{table:teaser4} shows the average time for each query of returning the different amounts of Nearest Neighbors (NN) on  Places205 and GIST1M, respectively. Linear Scan is accelerated by adopting the look-up tables~\cite{gordo2014asymmetric}. From the results, we can see that MIH and MIWQ are both faster than Linear Scan in all the cases. MIWQ is comparable or a litter inferior to MIH in the average time. From the tables, we can see that the query time of both MIH and MIWQ for 100-NN is larger than that for 1-NN. To return more neighbors about the query, more buckets need to be probed, resulting in a larger time cost.

\subsection{Case of Longer Binary Codes}
In the above experiments, we analyze the performance of our method for 32 bits and 64 bits, which are the commonly used length of binary codes for the hashing methods. In some situations, longer binary codes (such as 128 bits and 256 bits) are used to achieve higher search accuracy but with additional storage cost. Here, we analyze the performance of our method in the case of long binary codes.

Table~\ref{table:teaser5} shows the average time for the query on Places205. For 128 bits and 256 bits, MIWQ can still accelerate the search on the binary codes. By comparing Table~\ref{table:teaser5} to Table~\ref{table:teaser3}, the speed-up factors for 128 bit and 256 bits are smaller than the ones for 32 bits and 64 bits.

\begin{table*}[!ht]
	\caption{The average time for the query on Places205 with longer binary codes.}
	\begin{center}
		\begin{tabular}
            {|c|c|c|c|c|c|c|c|}\hline
          \multirow{3}{*}{bit} & \multirow{3}{*}{method} & \multicolumn{6}{c|}{speed-up factors for $K$-NN}        \\ \cline{3-8}
			                   &         &  \multicolumn{2}{c|}{1-NN} & \multicolumn{2}{c|}{10-NN} & \multicolumn{2}{c|}{100-NN} \\ \cline{3-8}
			                   &         & time(ms) 	 & speed-up factor    & time(ms)  & speed-up factor  & time(ms)	& speed-up factor  \\ \hline
         \multirow{3}{*}{128} &  Linear Scan  &  65.93	 & 1.0	       &  65.93	   & 1.0	   & 65.93	    & 1.0   \\ \cline{2-2}
                             &  MIH           &  2.08	 & 31.6        &  4.19	   & 15.7	   & 8.50	    & 7.7  \\ \cline{2-2}
                             &  $\bold{MIWQ}$          &  $\bold{3.13}$	 & $\bold{21.0}$	       &  $\bold{6.71}$	   & $\bold{9.8}$	   & $\bold{15.25}$	    & $\bold{4.3}$   \\ \cline{1-8}
         \multirow{3}{*}{256} &  Linear Scan  &  103.92  & 1.0	       &  103.92   & 1.0	   & 103.92	    & 1.0   \\ \cline{2-2}
                             &  MIH          &  5.68	 & 18.2	       &  11.80	   & 8.8	   & 21.22	    & 4.8  \\ \cline{2-2}
                             &  $\bold{MIWQ}$         &  $\bold{9.30}$	 & $\bold{11.1}$	       &  $\bold{21.97}$	   & $\bold{4.7}$	   & $\bold{39.28}$	    & $\bold{2.6}$  \\ \hline
		\end{tabular}
	\end{center}
	\label{table:teaser5}
\end{table*}

\subsection{Case of Larger Dataset}
Table~\ref{table:teaser6} shows the precision@$K$ results on SIFT1B. From the results, we can see that MIWQ still achieves better search accuracy than MIH and Baseline.

\begin{table}[!ht]

	\caption{The precision results for the query on SIFT1B.}
	\begin{center}
		\begin{tabular}
            {|c|c|c|c|c|}\hline
           \multirow{2}{*}{bit} & \multirow{2}{*}{method} & \multicolumn{3}{c|}{precision ($\%$)}                                 \\ \cline{3-5}
			                    &                         & 1-NN   & 10-NN    & 100-NN    \\ \hline
           \multirow{4}{*}{32} &  Baseline                & 1.61	   & 1.66	  & 1.77   \\ \cline{2-2}
                                &  MIH                    & 3.04	   & 2.04	  & 1.48   \\ \cline{2-2}
                                &  Linear Scan            & 3.25	   & 2.77	  & 2.66   \\ \cline{2-2}
                                & $\bold{MIWQ}$           & $\bf{3.25}$	   & $\bf{2.77}$	  & $\bf{2.66}$     \\ \hline
           \multirow{4}{*}{64} &  Baseline                & 7.62	   & 8.26	  & 9.07   \\ \cline{2-2}
                                &  MIH                    & 25.18	   & 20.22	  & 10.52  \\ \cline{2-2}
                                &  Linear Scan            & 26.77	   & 21.18	  & 14.13   \\ \cline{2-2}
                                & $\bold{MIWQ}$           & $\bf{26.77}$	   & $\bf{21.18}$	  & $\bf{14.13}$     \\ \hline
		\end{tabular}
	\end{center}
	\label{table:teaser6}

\end{table}

\begin{table*}[!ht]

	\caption{The average time for the query on SIFT1B.}
	\begin{center}
		\begin{tabular}
            {|c|c|c|c|c|c|c|c|}\hline
          \multirow{3}{*}{bit} & \multirow{3}{*}{method} & \multicolumn{6}{c|}{speed-up factors for $K$-NN}        \\ \cline{3-8}
			                   &         &  \multicolumn{2}{c|}{1-NN} & \multicolumn{2}{c|}{10-NN} & \multicolumn{2}{c|}{100-NN} \\ \cline{3-8}
			                   &         & time(ms) 	  & speed-up factor    & time(ms)  & speed-up factor  & time(ms)	& speed-up factor  \\ \hline
         \multirow{3}{*}{32} &  Linear Scan  &  22380.15 &	1.0	        & 22380.15	& 1.0	    & 22380.15	& 1.0   \\ \cline{2-2}
                             &  MIH           &  6.13	  & 3650.9	    & 6.14	    & 3644.9	& 6.38	    & 3507.8  \\ \cline{2-2}
                             &  $\bold{MIWQ}$          &  $\bold{6.15}$	  & $\bold{3639.0}$	    & $\bold{6.31}$	    & $\bold{3546.7}$	& $\bold{6.53}$	    & $\bold{3427.3}$  \\ \cline{1-8}
         \multirow{3}{*}{64} &  Linear Scan  &  43623.33 &	1.0	        & 43623.33	& 1.0	    & 43623.33	& 1.0   \\ \cline{2-2}
                             &  MIH          &  7.77	  & 5614.3	    & 11.42	    & 3819.9	& 22.96	    & 1899.9  \\ \cline{2-2}
                             &  $\bold{MIWQ}$         &  $\bold{7.78}$	  & $\bold{5607.1}$	    & $\bold{12.65}$	    & $\bold{3448.4}$	& $\bold{32.17}$	    & $\bold{1356.0}$ \\ \hline
		\end{tabular}
	\end{center}
	\label{table:teaser7}

\end{table*}

\begin{figure}[!ht]
\centering
\begin{tabular}{cc}

 \includegraphics[width=0.455\columnwidth]{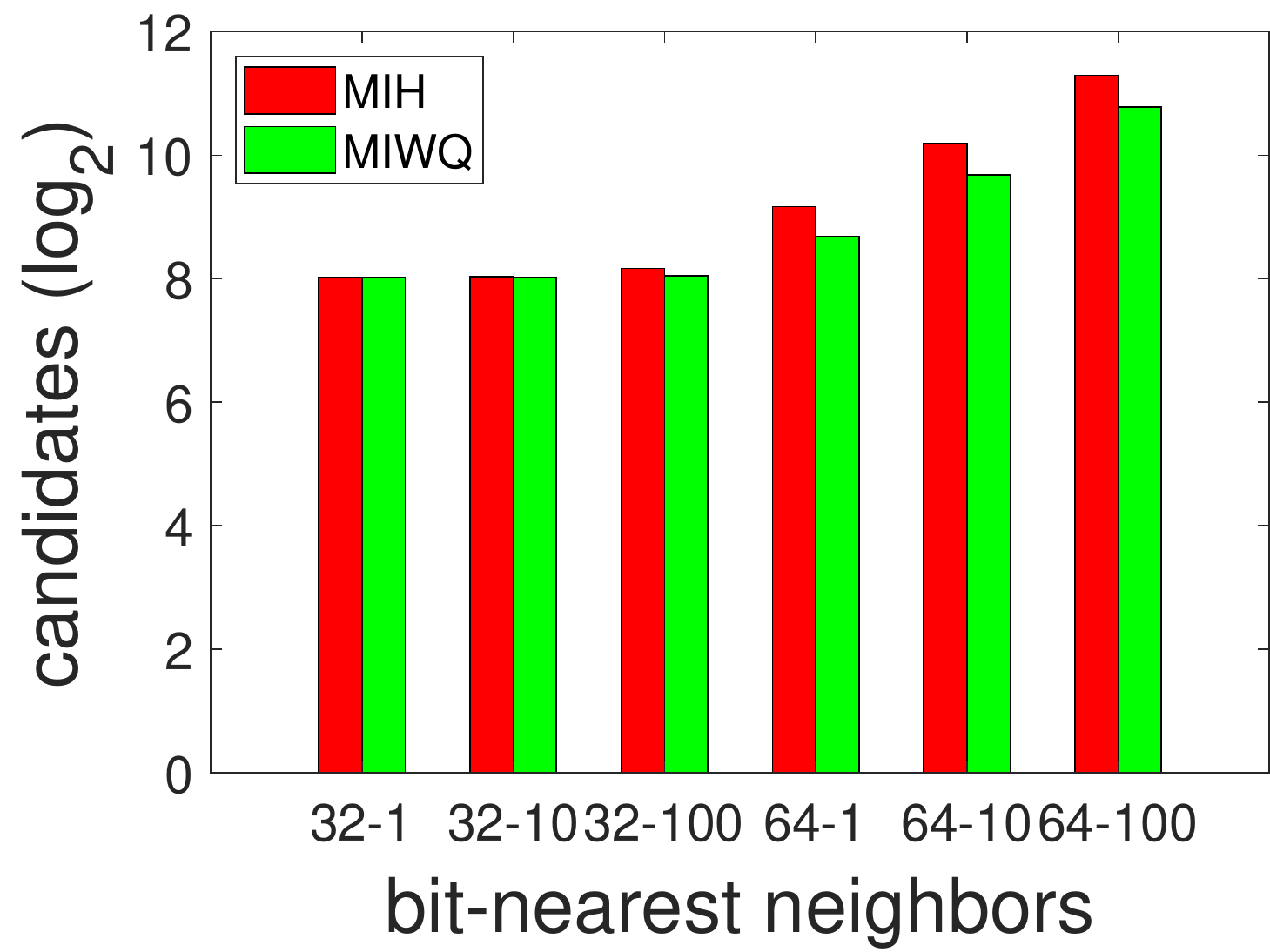}&
 \includegraphics[width=0.455\columnwidth]{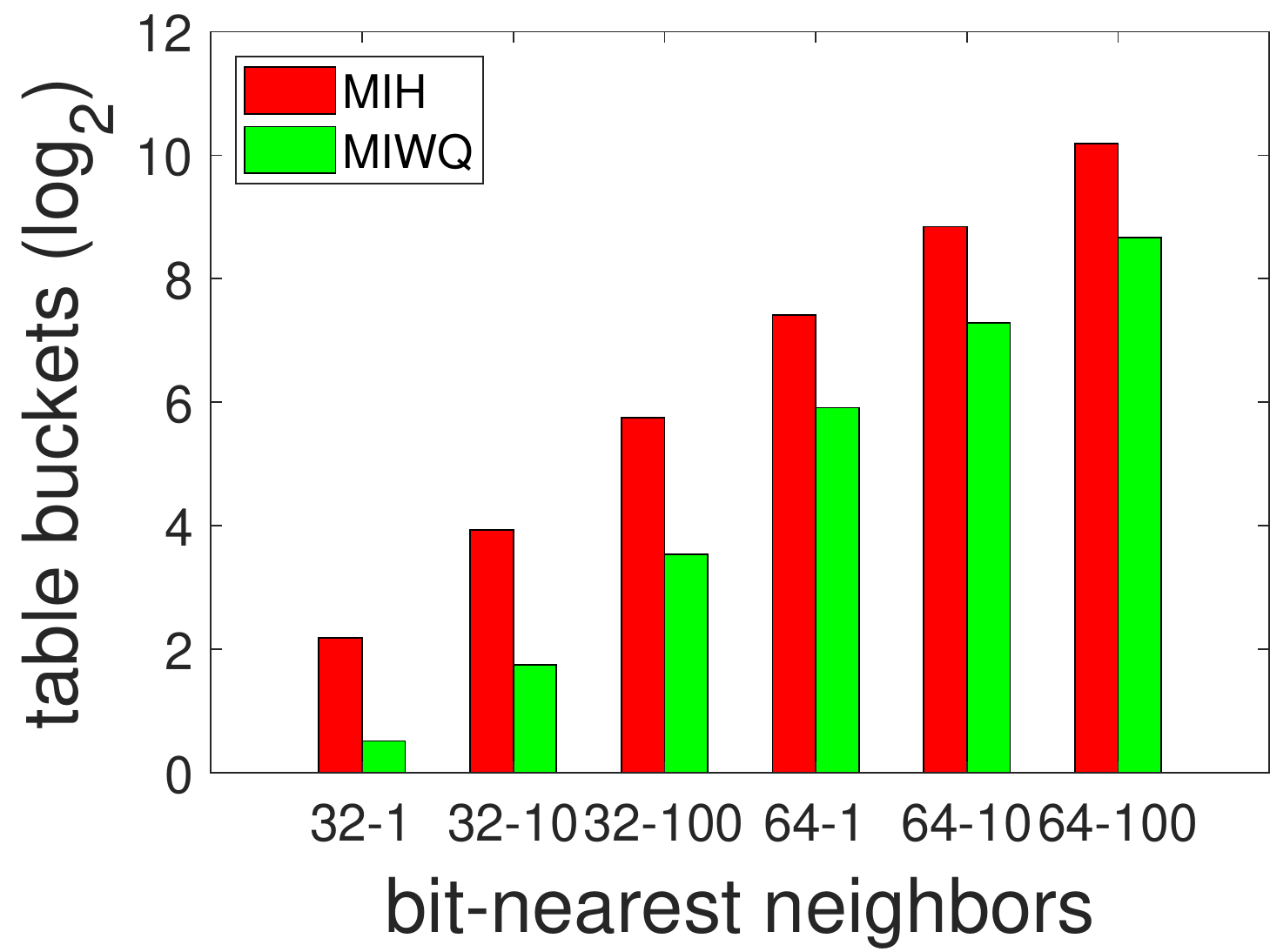}\\
{\small (a) candidates } &   {\small (b) buckets }

\end{tabular}

\caption{Comparison between MIWQ and MIH on SIFT1B.}
\label{fig:res18}

\end{figure}

Table~\ref{table:teaser7} shows the average time for the query on SIFT1B. From the results, we can see that MIH and MIWQ both have a large improvement on the speed compared to linear scan. MIWQ achieves almost the same time cost as MIH does for 32 bits and 64 bits. In the aspect of comparing candidates, MIWQ compares candidates by using weighted Hamming distance, while MIH compares candidates by using Hamming distance at first and then culls candidates by using weighted Hamming distance. Hence, MIH has a smaller time cost than our method. However, the factors to affect the search efficiency is not only the distance computation, but also the number of the candidates to be compared and the number of the table buckets to be probed. The number of candidates and the number of table buckets are shown in Fig.~\ref{fig:res18}. Since the average number of candidates and table buckets in our method are both smaller than those in MIH, our method can be almost as fast as MIH.

\subsection{PQTable}

Recently, PQTable~\cite{matsui2015pqtable} is proposed to perform an efficient search for Product Quantization (PQ)~\cite{jegou2011product} which is another encoding method. With some modifications, PQTable can be used for querying from the weighted binary codes. The binary codes are split into disjoint parts each of which consists of continuous 8-bit binary codes. Then, each part can be regarded as a codebook, and PQTable is applied.

Since PQTable and our method both can return the nearest weighted binary codes of the query accurately, we compare them with respect to the running time.
As PQTable is also based on the multi-index tables, to further explore the difference between our method and PQTable, our table merge algorithm is applied to PQTable and replaces the table merge algorithm of PQTable, which is dubbed as $\rm{PQTable_{max\_heap}}$. Table~\ref{table:teaser8} shows the time comparison between MIWQ, PQTable and $\rm{PQTable_{max\_heap}}$. According to the results, our table merging algorithm is faster than that of PQTable, which shows that our table merging algorithm can terminate the process by determining whether the $K$-NN of the query have been found faster than that of PQTable. Comparing MIWQ with $\rm{PQTable_{max\_heap}}$, with same table merging algorithm, MIWQ is faster than $\rm{PQTable_{max\_heap}}$, especially for the 32-bit case. As MIWQ and $\rm{PQTable_{max\_heap}}$ both perform the exact $K$-NN search, the order of table buckets to be probed is the same. The difference between them is the process to find the next hash table bucket. Since our method exploits the characteristics of the weighted binary codes, the bucket candidate space to traverse from our method is smaller than that of PQTable. Hence, our method can find the next smallest un-probed binary bucket faster than PQTable.

\begin{table}[!ht]

	\caption{Time cost ({\upshape ms}) on SIFT1B.}
	\begin{center}
		\begin{tabular}
            {|c|c|c|c|}\hline
            {bit-$K$ NN}     &     MIWQ    &    PQTable    &    $\rm{PQTable_{max\_heap}}$ \\ \hline
             32-1            &	6.15   &  51.31  &	50.44 \\ \hline
             32-10           &	6.31   &  51.22  &	50.48 \\ \hline
             32-100          &  6.53   &  51.43  &  50.55 \\ \hline
             64-1            &	7.78   &  57.26  &  54.53 \\ \hline
             64-10           &	12.65   &  64.55  &  61.21 \\ \hline
             64-100          &	32.17   &  92.11  &  85.33 \\ \hline

		\end{tabular}
	\end{center}
	\label{table:teaser8}

\end{table}

\section{Conclusion}
In this paper, a new querying method is proposed to return the nearest weighted binary codes of the query in a non-exhaustive way. The method consists of two algorithms, the table bucket finding algorithm and the table merging algorithm. The former one is designed to consecutively find the un-probed table buckets, and the latter one is developed to merge the candidates from each table. The experiments show that our method can produce the same querying results as linear scan does with a large time speed-up on the large-scale dataset which includes up to 1 billion data points.

\section{Acknowledge}
This work is supported by Shenzhen Municipal Development and Reform Commission (Disciplinary Development Program for Data Science and Intelligent Computing), Key-Area Research and Development Program of Guangdong Province ($\#$2019B010137001), NSFC-Shenzhen Robot Jointed Founding (U1613215).

\bibliographystyle{aaai}
\bibliography{egbib}

\begin{thebibliography}{}

\bibitem[\protect\citeauthoryear{Andoni and Indyk}{2006}]{andoni2006near}
Andoni, A., and Indyk, P.
\newblock 2006.
\newblock Near-optimal hashing algorithms for approximate nearest neighbor in
  high dimensions.
\newblock In {\em 47th Annual IEEE Symposium on Foundations of Computer
  Science},  459--468.

\bibitem[\protect\citeauthoryear{{Babenko} and {Lempitsky}}{2015}]{6915715}
{Babenko}, A., and {Lempitsky}, V.
\newblock 2015.
\newblock The inverted multi-index.
\newblock {\em IEEE TPAMI} 37(6):1247--1260.

\bibitem[\protect\citeauthoryear{{Balntas}, {Tang}, and
  {Mikolajczyk}}{2018}]{7882718}
{Balntas}, V.; {Tang}, L.; and {Mikolajczyk}, K.
\newblock 2018.
\newblock Binary online learned descriptors.
\newblock {\em IEEE TPAMI} 40(3):555--567.

\bibitem[\protect\citeauthoryear{Cakir \bgroup et al\mbox.\egroup
  }{2017}]{Cakir2017MIHash}
Cakir, F.; He, K.; Bargal, S.~A.; and Sclaroff, S.
\newblock 2017.
\newblock Mihash: Online hashing with mutual information.
\newblock In {\em ICCV},  437--445.

\bibitem[\protect\citeauthoryear{Duan \bgroup et al\mbox.\egroup
  }{2015}]{Duan2015Weighted}
Duan, L.~Y.; Lin, J.; Wang, Z.; Huang, T.; and Gao, W.
\newblock 2015.
\newblock Weighted component hashing of binary aggregated descriptors for fast
  visual search.
\newblock {\em IEEE TMM} 17(6):828--842.

\bibitem[\protect\citeauthoryear{Fan \bgroup et al\mbox.\egroup
  }{2013}]{Fan2013Learning}
Fan, B.; Kong, Q.; Yuan, X.; Wang, Z.; and Pan, C.
\newblock 2013.
\newblock Learning weighted hamming distance for binary descriptors.
\newblock In {\em ICASSP},  2395--2399.

\bibitem[\protect\citeauthoryear{Gong \bgroup et al\mbox.\egroup
  }{2013}]{gong2013iterative}
Gong, Y.; Lazebnik, S.; Gordo, A.; and Perronnin, F.
\newblock 2013.
\newblock Iterative quantization: A procrustean approach to learning binary
  codes for large-scale image retrieval.
\newblock {\em IEEE TPAMI} 35(12):2916--2929.

\bibitem[\protect\citeauthoryear{Gordo \bgroup et al\mbox.\egroup
  }{2014}]{gordo2014asymmetric}
Gordo, A.; Perronnin, F.; Gong, Y.; and Lazebnik, S.
\newblock 2014.
\newblock Asymmetric distances for binary embeddings.
\newblock {\em IEEE TPAMI} 36(1):33--47.

\bibitem[\protect\citeauthoryear{Huang, Wei, and Zhang}{2017}]{Huang2017RWBD}
Huang, Z.; Wei, Z.; and Zhang, G.
\newblock 2017.
\newblock Rwbd: Learning robust weighted binary descriptor for image matching.
\newblock {\em IEEE TCSVT} PP(99):1--1.

\bibitem[\protect\citeauthoryear{Jegou, Douze, and
  Schmid}{2011}]{jegou2011product}
Jegou, H.; Douze, M.; and Schmid, C.
\newblock 2011.
\newblock Product quantization for nearest neighbor search.
\newblock {\em IEEE TPAMI} 33(1):117--128.

\bibitem[\protect\citeauthoryear{Krizhevsky, Sutskever, and
  Hinton}{2012}]{Krizhevsky2012ImageNet}
Krizhevsky, A.; Sutskever, I.; and Hinton, G.~E.
\newblock 2012.
\newblock Imagenet classification with deep convolutional neural networks.
\newblock In {\em NeurIPS},  1097--1105.

\bibitem[\protect\citeauthoryear{{Leutenegger}, {Chli}, and
  {Siegwart}}{2011}]{6126542}
{Leutenegger}, S.; {Chli}, M.; and {Siegwart}, R.~Y.
\newblock 2011.
\newblock Brisk: Binary robust invariant scalable keypoints.
\newblock In {\em ICCV},  2548--2555.

\bibitem[\protect\citeauthoryear{Li \bgroup et al\mbox.\egroup
  }{2019}]{DBLP:conf/aaai/LiDWXL19}
Li, C.; Deng, C.; Wang, L.; Xie, D.; and Liu, X.
\newblock 2019.
\newblock Coupled cyclegan: Unsupervised hashing network for cross-modal
  retrieval.
\newblock In {\em AAAI},  176--183.

\bibitem[\protect\citeauthoryear{Lin \bgroup et al\mbox.\egroup
  }{2019}]{DBLP:conf/aaai/LinJLSWW19}
Lin, M.; Ji, R.; Liu, H.; Sun, X.; Wu, Y.; and Wu, Y.
\newblock 2019.
\newblock Towards optimal discrete online hashing with balanced similarity.
\newblock In {\em AAAI},  8722--8729.

\bibitem[\protect\citeauthoryear{Liu \bgroup et al\mbox.\egroup
  }{2014}]{liu2014discrete}
Liu, W.; Mu, C.; Kumar, S.; and Chang, S.-F.
\newblock 2014.
\newblock Discrete graph hashing.
\newblock In {\em NeurIPS},  3419--3427.

\bibitem[\protect\citeauthoryear{Liu \bgroup et al\mbox.\egroup
  }{2019}]{DBLP:journals/pami/LiuJWS19}
Liu, H.; Ji, R.; Wang, J.; and Shen, C.
\newblock 2019.
\newblock Ordinal constraint binary coding for approximate nearest neighbor
  search.
\newblock {\em IEEE TPAMI} 41(4):941--955.

\bibitem[\protect\citeauthoryear{Lowe}{2004}]{lowe2004distinctive}
Lowe, D.~G.
\newblock 2004.
\newblock Distinctive image features from scale-invariant keypoints.
\newblock {\em IJCV} 60(2):91--110.

\bibitem[\protect\citeauthoryear{Matsui, Yamasaki, and
  Aizawa}{2018}]{matsui2015pqtable}
Matsui, Y.; Yamasaki, T.; and Aizawa, K.
\newblock 2018.
\newblock Pqtable: Nonexhaustive fast search for product-quantized codes using
  hash tables.
\newblock {\em IEEE TMM} 20(7):1809--1822.

\bibitem[\protect\citeauthoryear{Norouzi, Punjani, and
  Fleet}{2014}]{norouzi2014fast}
Norouzi, M.; Punjani, A.; and Fleet, D.~J.
\newblock 2014.
\newblock Fast exact search in hamming space with multi-index hashing.
\newblock {\em IEEE TPAMI} 36(6):1107--1119.

\bibitem[\protect\citeauthoryear{Oliva and Torralba}{2001}]{oliva2001modeling}
Oliva, A., and Torralba, A.
\newblock 2001.
\newblock Modeling the shape of the scene: A holistic representation of the
  spatial envelope.
\newblock {\em IJCV} 42(3):145--175.

\bibitem[\protect\citeauthoryear{{Rublee} \bgroup et al\mbox.\egroup
  }{2011}]{6126544}
{Rublee}, E.; {Rabaud}, V.; {Konolige}, K.; and {Bradski}, G.
\newblock 2011.
\newblock Orb: An efficient alternative to sift or surf.
\newblock In {\em ICCV},  2564--2571.

\bibitem[\protect\citeauthoryear{Song \bgroup et al\mbox.\egroup
  }{2018}]{DBLP:conf/aaai/SongHGXHS18}
Song, J.; He, T.; Gao, L.; Xu, X.; Hanjalic, A.; and Shen, H.~T.
\newblock 2018.
\newblock Binary generative adversarial networks for image retrieval.
\newblock In {\em AAAI},  394--401.

\bibitem[\protect\citeauthoryear{Torralba, Fergus, and
  Freeman}{2008}]{torralba200880}
Torralba, A.; Fergus, R.; and Freeman, W.~T.
\newblock 2008.
\newblock 80 million tiny images: A large data set for nonparametric object and
  scene recognition.
\newblock {\em IEEE TPAMI} 30(11):1958--1970.

\bibitem[\protect\citeauthoryear{Wang \bgroup et al\mbox.\egroup
  }{2018}]{7915742}
Wang, J.; Zhang, T.; Song, J.; Sebe, N.; and Shen, H.~T.
\newblock 2018.
\newblock A survey on learning to hash.
\newblock {\em IEEE TPAMI} 40(4):769--790.

\bibitem[\protect\citeauthoryear{Weng \bgroup et al\mbox.\egroup
  }{2016}]{weng2016asymmetric}
Weng, Z.; Yao, W.; Sun, Z.; and Zhu, Y.
\newblock 2016.
\newblock Asymmetric distance for spherical hashing.
\newblock In {\em ICIP},  206--210.

\bibitem[\protect\citeauthoryear{Zhang \bgroup et al\mbox.\egroup
  }{2013}]{zhang2013binary}
Zhang, L.; Zhang, Y.; Tang, J.; Lu, K.; and Tian, Q.
\newblock 2013.
\newblock Binary code ranking with weighted hamming distance.
\newblock In {\em CVPR},  1586--1593.

\bibitem[\protect\citeauthoryear{Zhou \bgroup et al\mbox.\egroup
  }{2014}]{zhou2014learning}
Zhou, B.; Lapedriza, A.; Xiao, J.; Torralba, A.; and Oliva, A.
\newblock 2014.
\newblock Learning deep features for scene recognition using places database.
\newblock In {\em NeurIPS},  487--495.

\end{thebibliography}

\clearpage
\appendixpage

\section{Appendix A}
$\bold{Corollary}$ $\bold{1:}$ Every binary code can be generated by above two operations.

$\bold{Proof:}$ It can be proved by mathematical induction.

$\bf{Basis:}$ We have the binary code ${\bf{h}}_0$ = $\bf{h}$ which have no changed bits initially. Then by definition, ${\bf{\hat h}}_1$ is generated by changing the first bit of ${\bf{h}}_0$ according to the first operation. It is easy to find that every binary code ${\bf{h}}_1$ which have 1 changed bit can be generated from ${\bf{\hat h}}_1$ according to the second operation.

$\bf{Inductive}$ $\bold{step:}$ Assume every binary code ${\bf{h}}_z$ which has $z$ changed bits can be generated. For every binary code ${\bf{h}}_{z+1} = [h_1 \ldots \bar h_i \ldots \bar h_j \ldots h_b]$ which has $z+1$ changed bits, where the $i^{th}$ bit and the $j^{th}$ bit are the $z^{th}$ and $(z+1)^{th}$ changed bit, respectively. It can be generated by the second operation from another binary code ${\bf{\hat h}}_{z+1} = [h_1 \ldots \bar h_i \bar h_{i+1} \ldots h_b]$, where the changed status of the $j^{th}$ bit is moved to the $(i+1)^{th}$ bit. Then ${\bf{\hat h}}_{z+1}$ can be generated by the first operation from the binary code ${\bf{\tilde h}}_z = [h_1 \ldots \bar h_i h_{i+1} \ldots h_j \ldots h_b]$, where the $(i+1)^{th}$ bit is changed back to the previous status and ${\bf{\tilde h}}_z$ has $z$ changed bits. Thereby, every binary code ${\bf{h}}_{z+1}$ which has $z+1$ changed bits can be generated by two operations from $\bold{h}_z$.

\section{Appendix B}
$\bold{Theorem}$ $\bold{1:}$ The binary code that has the smallest sum of weights among the un-probed binary codes is always in the priority queue.

$\bold{Proof:}$ It can be proved by mathematical induction.

$\bf{Basis:}$ By definition, we have the binary code $\bf{h}$ which have no changed bits initially. It is the smallest among all the binary codes and is pushed into the priority queue.

$\bf{Inductive}$ $\bold{step:}$ Assume the top item ${\bf{h}}_a$ of the priority queue is the smallest among the current un-probed binary codes. When it is taken out, two new binary codes are constructed and pushed into the queue. Assume there is another binary code ${\bf{h}}_b$ which is not in the queue and is smallest among the current un-probed binary codes after ${\bf{h}}_a$ is taken out. According to $\bold{Corollary}$ $\bold{1}$, ${\bf{h}}_b$ can be directly generated from ${\bf{h}}_c$ by the above two operations. ${\bf{h}}_c$ is smaller than ${\bf{h}}_b$. Since ${\bf{h}}_b$ is smallest among the current un-probed binary codes, ${\bf{h}}_c$ should have been probed. If ${\bf{h}}_c$ is probed and ${\bf{h}}_b$ is generated from ${\bf{h}}_c$, ${\bf{h}}_b$ should have been pushed into the queue. Obviously, the assumption is invalid. So the binary code which has smallest sum of weights among the un-probed binary codes is always in the priority queue.

\section{Appendix C}
$\bold{Theorem}$ $\bold{2:}$ The binary codes that are found and stored in the max-heap have the smallest sum of weights among all binary codes.

$\bold{Proof:}$ This can be proved by contradiction. Assume there exists an identifier that is not found yet and its corresponding binary code is $\bold{\tilde r}$. Its sum of weights $g(\bold{\tilde r})$ (the definition of the function $g()$ is in Eqn.(7) in the manuscript) is smaller than that of the root node $\bold{r}$ in the max-heap ($i.e.$ $g(\bold{\tilde r}) < g(\bold{r})$). Since $g({\bf{r}}) \le S$ (the definition of $S$ is provided in Eqn.(8) in the manuscript) and $g({\bf{\tilde r}}) < g({\bf{r}}),$ $g({\bf{\tilde r}}) < S$. Then, at least one of the disjoint substrings of $\bold{\tilde r}$ is smaller than the top substring of the corresponding queue. This substring should have been taken out from the queue. Since the substring is taken out as a bucket to be probed, $\bold{\tilde r}$ should have been inserted into the max-heap. Obviously the assumption is invalid. Therefore, the binary codes that are found and stored in the max-heap have the smallest sum of weights among all binary codes.

\section{Appendix D}
To evaluate the efficiency and the accuracy of the querying methods on different weighted binary codes, the classical data-independent hashing algorithm Locality-Sensitive Hashing (LSH)~\cite{andoni2006near} and the classical data-dependent hashing algorithm Iterative Quantization (ITQ)~\cite{gong2013iterative} are used to map high-dimensional vectors into binary codes, and two weighted hashing method, Asymmetric Distance (Asym)~\cite{gordo2014asymmetric} and Weighted Hamming distance (Wh)~\cite{zhang2013binary}, are adopted to generate the weights for each bit of binary codes.

The average time for the query with different binary codes and different weights are shown from Table~\ref{table:teaser15} to Table~\ref{table:teaser18}. From the results, we can see that MIH and MIWQ are both faster than Linear Scan in all the cases. MIWQ is comparable or a litter inferior to MIH in the average time.

\begin{table*}[!ht]
	\caption{The average time for the query on Places205 with Asym.}
	\begin{center}
		\begin{tabular}
            {|c|c|c|c|c|c|c|c|c|}\hline
          \multirow{3}{*}{} & \multirow{3}{*}{bit} & \multirow{3}{*}{method} & \multicolumn{6}{c|}{speed-up factors for $K$-NN}        \\ \cline{4-9}
			          &               &         &  \multicolumn{2}{c|}{1-NN} & \multicolumn{2}{c|}{10-NN} & \multicolumn{2}{c|}{100-NN} \\ \cline{4-9}
			          &               &         & time(ms) 	 & speed-up factor   & time(ms)  & speed-up factor & time(ms)	& speed-up factor  \\ \hline
        \multirow{6}{*}{LSH}& \multirow{3}{*}{32} &  Linear Scan  &  23.06 & 1.0	& 23.06	 &  1.0	  &   23.06	 &  1.0   \\ \cline{3-3}
                            &                     &  MIH          &  0.11  & 209.6	& 0.21	 &  109.8 &   0.52	 &  44.3  \\ \cline{3-3}
                            &                     &  $\bf{MIWQ}$         &  $\bf{0.11}$  & $\bf{209.6}$	& $\bf{0.21}$	 & $\bf{109.8}$  &	  $\bf{0.55}$	 & $\bf{41.9}$   \\ \cline{2-9}
                            & \multirow{3}{*}{64} &  Linear Scan  & 33.83  & 1.0	& 33.83	 & 1.0	  &   33.83	 & 1.0   \\ \cline{3-3}
                            &                     &  MIH          &  0.51  & 66.3	& 1.03	 & 32.84  &   2.4	 & 14.0  \\ \cline{3-3}
                            &                     &  $\bf{MIWQ}$         &  $\bf{0.63}$  & $\bf{53.6}$	& $\bf{1.42}$	 & $\bf{23.82}$  &   $\bf{3.65}$	 & $\bf{9.2}$  \\ \hline
        \multirow{6}{*}{ITQ}& \multirow{3}{*}{32} &  Linear Scan  &  18.81 & 1.0	& 18.81	 & 1.0	  &   18.81	 & 1.0   \\ \cline{3-3}
                            &                     &  MIH          &  0.08  & 235.1	& 0.11	 & 171.0  &   0.23   & 81.7  \\ \cline{3-3}
                            &                     &  $\bf{MIWQ}$         &  $\bf{0.08}$  & $\bf{235.1}$	& $\bf{0.12}$	 & $\bf{156.7}$  &   $\bf{0.28}$	 & $\bf{67.1}$ \\ \cline{2-9}
                            & \multirow{3}{*}{64} &  Linear Scan  &  29.9  & 1.0	& 29.9	 & 1.0	  &   29.9	 & 1.0   \\ \cline{3-3}
                            &                     &  MIH          &  0.28  & 106.7	& 0.52	 & 57.5   &   1.1    & 27.1  \\ \cline{3-3}
                            &                     &  $\bf{MIWQ}$         &  $\bf{0.39}$  & $\bf{76.6}$	& $\bf{0.8}$	 & $\bf{37.3}$   &   $\bf{1.91}$	 & $\bf{15.6}$  \\ \hline
		\end{tabular}
	\end{center}
	\label{table:teaser15}
\vspace{-5mm}
\end{table*}

\begin{table*}[!ht]
	\caption{The average time for the query on Places205 with Wh.}
	\begin{center}
		\begin{tabular}
            {|c|c|c|c|c|c|c|c|c|}\hline
          \multirow{3}{*}{} & \multirow{3}{*}{bit} & \multirow{3}{*}{method} & \multicolumn{6}{c|}{speed-up factors for $K$-NN}        \\ \cline{4-9}
			          &               &         &  \multicolumn{2}{c|}{1-NN} & \multicolumn{2}{c|}{10-NN} & \multicolumn{2}{c|}{100-NN} \\ \cline{4-9}
			          &               &         & time(ms) 	 & speed-up factor    & time(ms)  & speed-up factor  & time(ms)	& speed-up factor  \\ \hline
        \multirow{6}{*}{LSH}& \multirow{3}{*}{32} &  Linear Scan  &  24.43 & 1.0	& 24.43	 & 1.0	  & 24.43	 & 1.0   \\ \cline{3-3}
                            &                     &  MIH          &  0.11  & 222.0	& 0.21	 & 116.3  & 0.52	 & 46.9  \\ \cline{3-3}
                            &                     &  $\bf{MIWQ}$         &  $\bf{0.11}$  & $\bf{222.0}$	& $\bf{0.21}$	 & $\bf{116.3}$  & $\bf{0.54}$	 & $\bf{45.2}$   \\ \cline{2-9}
                            & \multirow{3}{*}{64} &  Linear Scan  &  46.23 & 1.0	& 46.23	 & 1.0	  & 46.23	 & 1.0   \\ \cline{3-3}
                            &                     &  MIH          &  0.51  & 90.6	& 1.03	 & 44.8   &	2.4	     & 19.2  \\ \cline{3-3}
                            &                     &  $\bf{MIWQ}$         &  $\bf{0.62}$  & $\bf{74.5}$	& $\bf{1.39}$	 & $\bf{33.2}$	  & $\bf{3.58}$	 & $\bf{12.9}$  \\ \hline
        \multirow{6}{*}{ITQ}& \multirow{3}{*}{32} &  Linear Scan  &  24.71 & 1.0	& 24.71	 & 1.0	  & 24.71	 & 1.0   \\ \cline{3-3}
                            &                     &  MIH          &  0.08  & 308.8	& 0.11	 & 224.6  & 0.23     & 107.4  \\ \cline{3-3}
                            &                     &  $\bf{MIWQ}$         &   $\bf{0.08}$ & $\bf{308.8}$	& $\bf{0.12}$	 & $\bf{205.9}$  & $\bf{0.3}$	     & $\bf{82.3}$ \\ \cline{2-9}
                            & \multirow{3}{*}{64} &  Linear Scan  &  46.21 & 1.0	& 46.21	 & 1.0	  & 46.21	 & 1.0   \\ \cline{3-3}
                            &                     &  MIH          &  0.28  & 165.0	& 0.52	 & 88.8	  & 1.1	     & 42.0  \\ \cline{3-3}
                            &                     &  $\bf{MIWQ}$        &  $\bf{0.41}$  & $\bf{112.7}$	& $\bf{0.86}$	 & $\bf{53.7}$   &	$\bf{2.05}$     & $\bf{22.5}$  \\ \hline
		\end{tabular}
	\end{center}
	\label{table:teaser16}
\vspace{-5mm}
\end{table*}

\begin{table*}[!ht]
	\caption{The average time for the query on GIST1M with Asym.}
	\begin{center}
		\begin{tabular}
            {|c|c|c|c|c|c|c|c|c|}\hline
          \multirow{3}{*}{} & \multirow{3}{*}{bit} & \multirow{3}{*}{method} & \multicolumn{6}{c|}{speed-up factors for $K$-NN}        \\ \cline{4-9}
			          &               &         &  \multicolumn{2}{c|}{1-NN} & \multicolumn{2}{c|}{10-NN} & \multicolumn{2}{c|}{100-NN} \\ \cline{4-9}
			          &               &         & time(ms) 	 & speed-up factor    & time(ms)  & speed-up factor  & time(ms)	& speed-up factor  \\ \hline
        \multirow{6}{*}{LSH}& \multirow{3}{*}{32} &  Linear Scan  &  22.17 & 1.0	& 22.17	 & 1.0	  &   22.17	 & 1.0   \\ \cline{3-3}
                            &                     &  MIH          &  0.1   & 221.7	& 0.16	 & 138.5  &   0.37	 & 59.9  \\ \cline{3-3}
                            &                     &  $\bf{MIWQ}$         &  $\bf{0.12}$  & $\bf{184.7}$	& $\bf{0.25}$	 & $\bf{88.6}$	  &   $\bf{0.77}$	 & $\bf{28.7}$   \\ \cline{2-9}
                            & \multirow{3}{*}{64} &  Linear Scan  & 39.36  & 1.0	& 39.36	 & 1.0	  &   39.36	 & 1.0   \\ \cline{3-3}
                            &                     &  MIH          &  0.87  & 45.2	& 1.65	 & 23.8	  &   3.15	 & 12.4  \\ \cline{3-3}
                            &                     &  $\bf{MIWQ}$         &  $\bf{1.84}$  & $\bf{21.3}$	& $\bf{3.9}$	 & $\bf{10.0}$	  &   $\bf{8.17}$	 & $\bf{4.8}$  \\ \hline
        \multirow{6}{*}{ITQ}& \multirow{3}{*}{32} &  Linear Scan  &  22.09 & 1.0	& 22.09	 & 1.0	  &   22.09	 & 1.0   \\ \cline{3-3}
                            &                     &  MIH          &  0.16  & 138.0	& 0.3	 & 73.6	  &   0.69	 & 32.0  \\ \cline{3-3}
                            &                     &  $\bf{MIWQ}$         &  $\bf{0.17}$  & $\bf{129.9}$	& $\bf{0.33}$	 & $\bf{66.9}$	  &   $\bf{0.84}$	 & $\bf{26.2}$ \\ \cline{2-9}
                            & \multirow{3}{*}{64} &  Linear Scan  &  38.71 & 1.0	& 38.71	 & 1.0	  &   38.71	 & 1.0  \\ \cline{3-3}
                            &                     &  MIH          &  1.35  & 28.6	& 2.6	 & 14.8	  &   5.04	 & 7.6  \\ \cline{3-3}
                            &                     &  $\bf{MIWQ}$         &  $\bf{2.14}$  & $\bf{18.0}$	& $\bf{4.55}$	 & $\bf{8.5}$	  &   $\bf{9.66}$	 & $\bf{4.0}$  \\ \hline
		\end{tabular}
	\end{center}
	\label{table:teaser17}
\vspace{-5mm}
\end{table*}

\begin{table*}[!ht]
	\caption{The average time for the query on GIST1M with Wh.}
	\begin{center}
		\begin{tabular}
            {|c|c|c|c|c|c|c|c|c|}\hline
          \multirow{3}{*}{} & \multirow{3}{*}{bit} & \multirow{3}{*}{method} & \multicolumn{6}{c|}{speed-up factors for $K$-NN}        \\ \cline{4-9}
			          &               &         &  \multicolumn{2}{c|}{1-NN} & \multicolumn{2}{c|}{10-NN} & \multicolumn{2}{c|}{100-NN} \\ \cline{4-9}
			          &               &         & time(ms) 	 & speed-up factor    & time(ms)  & speed-up factor  & time(ms)	& speed-up factor  \\ \hline
        \multirow{6}{*}{LSH}& \multirow{3}{*}{32} &  Linear Scan  &  23.93 & 1.0	& 23.93	 & 1.0	  &  23.93	 & 1.0   \\ \cline{3-3}
                            &                     &  MIH          &  0.1   & 239.3	& 0.16	 & 149.5  &  0.37	 & 64.6  \\ \cline{3-3}
                            &                     &  $\bf{MIWQ}$         &  $\bf{0.12}$  & $\bf{199.4}$	& $\bf{0.26}$	 & $\bf{92.0}$	  &  $\bf{0.78}$	 & $\bf{30.6}$   \\ \cline{2-9}
                            & \multirow{3}{*}{64} &  Linear Scan  &  38.52 & 1.0	& 38.52	 & 1.0	  &  38.52	 & 1.0   \\ \cline{3-3}
                            &                     &  MIH          &  0.87  & 44.2	& 1.65	 & 23.3	  &  3.15	 & 12.2  \\ \cline{3-3}
                            &                     &  $\bf{MIWQ}$         &  $\bf{1.92}$  & $\bf{20.0}$	& $\bf{4.06}$	 & $\bf{9.4}$	  &  $\bf{8.52}$	 & $\bf{4.5}$  \\ \hline
        \multirow{6}{*}{ITQ}& \multirow{3}{*}{32} &  Linear Scan  &  22.62 & 1.0	& 22.62  & 1.0	  &  22.62	 & 1.0   \\ \cline{3-3}
                            &                     &  MIH          &  0.16  & 141.3	& 0.3	 & 75.4   &	 0.69	 & 32.7  \\ \cline{3-3}
                            &                     &  $\bf{MIWQ}$         &  $\bf{0.18}$  & $\bf{125.6}$	& $\bf{0.31}$	 & $\bf{72.9}$	  &  $\bf{0.89}$	 & $\bf{25.4}$ \\ \cline{2-9}
                            & \multirow{3}{*}{64} &  Linear Scan  &  43.72 & 1.0	& 43.72	 & 1.0	  &  43.72	 & 1.0  \\ \cline{3-3}
                            &                     &  MIH          &  1.35  & 32.3	& 2.6	 & 16.8	  &  5.04	 & 8.6  \\ \cline{3-3}
                            &                     &  $\bf{MIWQ}$         &  $\bf{2.17}$  & $\bf{20.1}$	& $\bf{4.82}$	 & $\bf{9.0}$	  &  $\bf{9.91}$	 & $\bf{4.4}$ \\ \hline
		\end{tabular}
	\end{center}
	\label{table:teaser18}
\vspace{-5mm}
\end{table*}

The precision results of the querying methods with different binary codes and different weights are shown from Table~\ref{table:teaser31} to Table~\ref{table:teaser34}. According to the results, MIH can achieve higher search accuracy than Baseline, but is inferior to MIWQ. As MIH cannot return the $K$ nearest weighted binary codes accurately, MIH is inferior to MIWQ. Since MIWQ achieves the same search accuracy as Linear Scan, it shows that MIWQ can return the $K$ nearest weighted binary codes of the query accurately.

\begin{table*}[!ht]
	\caption{The precision results on Places205 with binary codes generated by LSH.}
	\begin{center}
		\begin{tabular}
            {|c|c|c|c|c|c|c|c|}\hline
           \multirow{3}{*}{bit} & \multirow{3}{*}{method} & \multicolumn{6}{c|}{precision ($\%$)}                                 \\ \cline{3-8}
			                    &                         & \multicolumn{3}{c|}{Asym}      & \multicolumn{3}{c|}{Wh} \\ \cline{3-8}
			                    &                         & 1-NN   & 10-NN    & 100-NN  & 1-NN   & 10-NN    & 100-NN  \\ \hline
           \multirow{4}{*}{32}  &  Baseline               & 21.56  & 22.23	  &  19.83	& 21.56  & 22.23	  & 19.83   \\ \cline{2-2}
                                &  MIH                    & 22.81  & 23.20	  &  20.86	& 22.70  & 23.22	  & 20.90   \\ \cline{2-2}
                                &  Linear Scan            & 25.29  & 24.97	  &  22.64	& 25.58  & 24.98	  & 22.71   \\ \cline{2-2}
                                & $\bold{MIWQ}$           & $\bf{25.29}$  & $\bf{24.97}$&  $\bf{22.64}$	 & $\bf{25.58}$  & $\bf{24.98}$	  & $\bf{22.71}$   \\ \hline
           \multirow{4}{*}{64}  &  Baseline               & 31.17  & 29.71	  &  26.67	& 31.17  & 29.71	  & 26.67   \\ \cline{2-2}
                                &  MIH                    & 31.41  & 30.27	  &  27.46	& 31.31  & 30.31	  & 27.49   \\ \cline{2-2}
                                &  Linear Scan            & 34.46  & 31.96	  &  28.72	& 34.34  & 31.87	  & 28.88   \\ \cline{2-2}
                                & $\bold{MIWQ}$           & $\bf{34.46}$  &	$\bf{31.96}$  &	$\bf{28.72}$   &	$\bf{34.34}$  &	$\bf{31.87}$  &	$\bf{28.88}$     \\ \hline
		\end{tabular}
	\end{center}
	\label{table:teaser31}
\vspace{-5mm}
\end{table*}

\begin{table*}[!ht]
	\caption{The precision results on Places205 with binary codes generated by ITQ.}
	\begin{center}
		\begin{tabular}
            {|c|c|c|c|c|c|c|c|}\hline
           \multirow{3}{*}{bit} & \multirow{3}{*}{method} & \multicolumn{6}{c|}{precision ($\%$)}                                 \\ \cline{3-8}
			                    &                         & \multicolumn{3}{c|}{Asym}      & \multicolumn{3}{c|}{Wh} \\ \cline{3-8}
			                    &                         & 1-NN   & 10-NN    & 100-NN    & 1-NN   & 10-NN    & 100-NN  \\ \hline
           \multirow{4}{*}{32}  &  Baseline               & 22.52  & 23.85	  & 22.40	 & 22.52  & 23.85	  & 22.40   \\ \cline{2-2}
                                &  MIH                    & 23.32  & 24.86	  &  23.12	 & 23.41  & 24.89	  & 23.12   \\ \cline{2-2}
                                &  Linear Scan            & 24.96  & 26.19	  &  24.42	 & 24.90  & 26.22	  & 24.41   \\ \cline{2-2}
                                & $\bold{MIWQ}$           & $\bf{24.96}$  &	$\bf{26.19}$  &	$\bf{24.42}$  &	$\bf{24.90}$  &	$\bf{26.22}$  &	$\bf{24.41}$     \\ \hline
           \multirow{4}{*}{64}  &  Baseline               & 32.76  & 31.53	  &  28.91	& 32.76  & 31.53	  & 28.91   \\ \cline{2-2}
                                &  MIH                    & 33.34  & 32.14	  &  29.53  & 36.68  & 35.63	  & 32.66   \\ \cline{2-2}
                                &  Linear Scan            & 35.02  & 33.50	  &  30.69  & 38.81  & 36.81	  & 33.55   \\ \cline{2-2}
                                & $\bold{MIWQ}$           & $\bf{35.02}$  &	$\bf{33.50}$  &	$\bf{30.69}$  &	$\bf{38.81}$  &	$\bf{36.81}$  &	$\bf{33.55}$       \\ \hline
		\end{tabular}
	\end{center}
	\label{table:teaser32}
\vspace{-5mm}
\end{table*}

\begin{table*}[!ht]
	\caption{The precision results on GIST1M with binary codes generated by LSH.}
	\begin{center}
		\begin{tabular}
            {|c|c|c|c|c|c|c|c|}\hline
           \multirow{3}{*}{bit} & \multirow{3}{*}{method} & \multicolumn{6}{c|}{precision ($\%$)}                                 \\ \cline{3-8}
			                    &                         & \multicolumn{3}{c|}{Asym}   & \multicolumn{3}{c|}{Wh} \\ \cline{3-8}
			                    &                         & 1-NN   & 10-NN    & 100-NN  & 1-NN   & 10-NN    & 100-NN  \\ \hline
           \multirow{4}{*}{32}  &  Baseline               & 11.90  & 7.57	  &  5.11	& 11.90  & 7.57	  & 5.11   \\ \cline{2-2}
                                &  MIH                    & 12.10  & 8.46	  &  5.92   & 11.80  & 8.12	  & 5.74   \\ \cline{2-2}
                                &  Linear Scan            & 13.20  & 9.94     &	 7.31   & 13.50  & 9.75     &	7.04   \\ \cline{2-2}
                                & $\bold{MIWQ}$           & $\bf{13.20}$  &	$\bf{9.94}$  &	$\bf{7.31}$  &	$\bf{13.50}$  &	$\bf{9.75}$  &	$\bf{7.04}$       \\ \hline
           \multirow{4}{*}{64}  &  Baseline               & 21.10  & 14.47    &	 10.16	& 21.10  & 14.47	  & 10.16   \\ \cline{2-2}
                                &  MIH                    & 21.50  & 16.11    &	 11.55	& 21.70  & 15.63    &	11.26   \\ \cline{2-2}
                                &  Linear Scan            & 24.90  & 19.14    &  13.65  & 23.90  & 18.30    & 13.36   \\ \cline{2-2}
                                & $\bold{MIWQ}$           & $\bf{24.90}$   & $\bf{19.14}$     &  $\bf{13.65}$     & $\bf{23.90}$   & $\bf{18.30}$     & $\bf{13.36}$        \\ \hline
		\end{tabular}
	\end{center}
	\label{table:teaser33}
\vspace{-5mm}
\end{table*}

\begin{table*}[!ht]
	\caption{The precision results on GIST1M with binary codes generated by ITQ.}
	\begin{center}
		\begin{tabular}
            {|c|c|c|c|c|c|c|c|}\hline
           \multirow{3}{*}{bit} & \multirow{3}{*}{method} & \multicolumn{6}{c|}{precision ($\%$)}                                 \\ \cline{3-8}
			                    &                         & \multicolumn{3}{c|}{Asym}    & \multicolumn{3}{c|}{Wh} \\ \cline{3-8}
			                    &                         & 1-NN   & 10-NN    & 100-NN  & 1-NN   & 10-NN    & 100-NN  \\ \hline
           \multirow{4}{*}{32}  &  Baseline               & 30.80  & 21.58    &	14.68   & 30.80  & 21.58    &	14.68   \\ \cline{2-2}
                                &  MIH                    & 29.50  & 23.29    &	16.76	& 30.50  & 22.58	  & 16.35   \\ \cline{2-2}
                                &  Linear Scan            & 33.80  & 26.69    &	19.70    &	34.2   & 26.01    &	19.06   \\ \cline{2-2}
                                & $\bold{MIWQ}$           & $\bf{33.80}$  & $\bf{26.69}$    &	$\bf{19.70}$    &	$\bf{34.2}$   & $\bf{26.01}$    &	$\bf{19.06}$       \\ \hline
           \multirow{4}{*}{64}  &  Baseline               & 39.60  & 30.15    &	 20.71	&	39.60  & 30.15    &	20.71   \\ \cline{2-2}
                                &  MIH                    & 40.40  & 33.69    &  24.02	& 39.80  & 32.13	  & 22.92   \\ \cline{2-2}
                                &  Linear Scan            & 47.00  & 38.40    &	 27.41	& 45.50  & 36.29	  & 26.43   \\ \cline{2-2}
                                & $\bold{MIWQ}$           & $\bf{47.00}$  & $\bf{38.40}$    &	 $\bf{27.41}$ & $\bf{45.50}$  & $\bf{36.29}$	  & $\bf{26.43}$   \\ \hline
		\end{tabular}
	\end{center}
	\label{table:teaser34}
\vspace{-5mm}
\end{table*}

\end{document}